\documentclass{article}
\PassOptionsToPackage{numbers, compress}{natbib}
\usepackage[preprint]{neurips_2024}

\usepackage[utf8]{inputenc} 
\usepackage[T1]{fontenc}    
\usepackage{hyperref}       
\usepackage{url}            
\usepackage{booktabs}       
\usepackage{amsfonts}       
\usepackage{nicefrac}       
\usepackage{microtype}      
\usepackage{xcolor}         
\usepackage{natbib}         
\usepackage{comment}
\usepackage{hyperref}
\usepackage{url}
\usepackage{graphicx}
\usepackage{multirow}
\usepackage{colortbl}

\title{SAFE setup for generative molecular design}

\author{
Yassir El Mesbahi \\
Valence Labs \\
\texttt{yassir@valencelabs.com}
\And
Emmanuel Noutahi \\
Valence Labs \\
\texttt{emmanuel@valencelabs.com}
}

\begin{document}
\maketitle
\begin{abstract}
SMILES-based molecular generative models have been pivotal in drug design but face challenges in fragment-constrained tasks. To address this, the Sequential Attachment-based Fragment Embedding (SAFE) representation was recently introduced as an alternative that streamlines those tasks. In this study, we investigate the optimal setups for training SAFE generative models, focusing on dataset size, data augmentation through randomization, model architecture, and bond disconnection algorithms. We found that larger, more diverse datasets improve performance, with the LLaMA architecture using Rotary Positional Embedding proving most robust. SAFE-based models also consistently outperform SMILES-based approaches in scaffold decoration and linker design, particularly with BRICS decomposition yielding the best results. These insights highlight key factors that significantly impact the efficacy of SAFE-based generative models.
\end{abstract}




\section{Introduction}

Molecular design is a fundamental task in computational drug discovery, aimed at constructing molecules with desired properties. Recently, deep generative models have emerged as valuable tools for efficiently exploring chemical space and designing novel molecules \citep{gomez2018automatic, merk2018novo, domenico2020novo, ozturk2020exploring, horwood2020molecular}. Among these, Chemical Language Models (CLMs) which use string-based molecular representations, typically the Simplified Molecular Input Line Entry System (SMILES) \citep{weininger1988smiles}, have shown robust performance by adapting neural architectures from natural language processing to generate molecular strings \citep{grisoni2023chemical}.

In practical drug design, it is often crucial to either preserve core 'scaffolds' while experimenting with various decoration groups or link several molecular fragments. This is essential for optimizing molecular properties, analyzing structure-activity relationships, and generating intellectual property. However, traditional CLMs require complex adaptations, including new architectures, retraining, custom data processing and sampling algorithms, to handle fragment-constrained tasks.

To address these challenges, the Sequential Attachment-based Fragment Embedding (SAFE) representation was introduced~\citep{noutahi2024gotta}. SAFE simplifies molecular design by treating molecules as unordered sequences of fragment blocks, transforming tasks such as \textit{de novo} generation, fragment linking, and scaffold decoration into simple sequence completion problems, all while remaining compatible with existing SMILES parsers. 

Building on the original SAFE work and the generative SAFE-GPT-2 model it introduced, this study investigates how different experimental design choices impact the performance of SAFE-based generative models. Specifically, we assess:

\begin{itemize}
    \item The influence of bond disconnection algorithms on generative outcomes.
    \item The impact of dataset size on model performance.
    \item The performance of various neural architectures.
    \item The effects of data augmentation via SAFE fragment randomization.
\end{itemize}

Our goal is to identify the experimental conditions that maximize the generative capabilities of SAFE-based models, thereby providing a comprehensive understanding of the SAFE representation's strengths and limitations in molecular design.

\section{Related Works}
\subsection{Generative Chemical Language Models}

Sequence-based generative models, particularly those using molecular line notations like SMILES and SELFIES~\cite{krenn2022selfies}, have grown in popularity in molecular design. These autoregressive models, leveraging advances in natural language processing (e.g., RNNs, transformers, and State Space Models), have demonstrated strong performance in both \textit{de novo} generation and goal-directed molecule design \citep{moret2023leveraging, ozccelik2024chemical, grisoni2023chemical}.  While SELFIES ensures robustness and validity, it comes at the cost of simplicity, interpretability, and as recent studies suggest, exploration and generalization capabilities   compared to SMILES \citep{skinnider2024invalid}. Additionally, SMILES, with its non-injective nature, allows multiple valid strings for the same molecular graph. This feature, known as SMILES randomization, can be used as data augmentation strategy to enhance model robustness and generalization, especially in data-scarce settings \citep{bjerrum2017smiles, skinnider2021chemical, arus2019randomized, moret2020generative}.

\subsection{Fragment-Constrained Generation with CLMs}
The need for fragment-constrained design in drug discovery has driven various adaptations of CLMs. \citet{arus2020smiles} proposed an encoder-decoder architecture that frames the task as sequence translation.  It requires slicing the molecular dataset into scaffolds and attachment groups before training.  Initially, this method was used to translate input scaffolds into output decorations, and was later extended in LibINVENT \citep{fialkova2021libinvent} to generate novel scaffold-constrained molecules that follow bespoke chemical reaction rules. Building on this framework, LinkINVENT \citep{guo2023link} and SyntaLinker \citep{yang2020syntalinker} further refined the encoder-decoder architecture. LinkINVENT reverses the translation task by taking a pair of molecular fragments to predict a linker, while SyntaLinker uses a conditional transformer model to translate fragments into fully linked molecules. However, these methods require multiple task-specific architectures, making them less practical due to extensive retraining for each unique chemical design challenge.

More flexible approaches, such as SAMOA \citep{langevin2020scaffold} and PromptSMILES \citep{thomas2024promptsmiles}, extend existing SMILES-based CLMs to scaffold decoration and linker design without requiring retraining or custom datasets. SAMOA allows free sampling at attachment/linking points but lacks guarantees for validity or scaffold constraints, while PromptSMILES rearranges SMILES strings iteratively to position attachment points for completion, and can be further complemented by  reinforcement learning for fine-tuning towards specific goals.

In contrast, SAFE \citep{noutahi2024gotta}, a sub-grammar of SMILES that reorganizes SMILES strings as unordered sequences of interconnected fragment blocks, simplifies fragment-constrained generation into a sequence completion task. SAFE eliminates the need for complex decoding schemes and has proven effective in both \textit{de novo} and fragment-constrained tasks using a pretrained GPT-2 model. Furthermore, the approach can benefit not only from SMILES randomization but also from a novel form of data augmentation (SAFE randomization), where the order of fragment blocks is randomized to further diversify training data.

\section{Experimental Setup}
\subsection{Building Chemical Language Models}
\subsubsection{Dataset}
We conducted our experiments using the MOSES benchmark dataset \citep{polykovskiy2020molecular} provided by TDC \citep{Huang2021tdc}, consisting of $\sim$1.3M training molecules, $\sim$193k validation, and $\sim$387k test molecules. The dataset was curated from the ZINC clean lead collection by removing molecules with charged atoms; atoms besides C, N, S, O, F, Cl, Br, H; large cycles, and those failing custom medicinal chemistry filters. For studying the impact of training set size, we downsampled the training data into subsets of 10k and 100k molecules, referred to as MOSES-10k and MOSES-100k, respectively, while the original dataset will be referred to as MOSES-Full.

For each dataset, we converted SMILES into canonical SAFE format using five fragmentation algorithms: Hussain-Rea (HR) \citep{hussain2010computationally}, BRICS \citep{degen2008art}, RECAP \citep{lewell1998recap}, RDKit's MMPA bond rules (MMPA) \citep{dalke2018mmpdb}, and Rotatable bonds (ROTATABLE) \citep{scharfer2013torsion}. In rare cases where fragmentation failed, resulting molecules were discarded. To evaluate data augmentation effects, we generated a 5x augmented version of MOSES-100k via SAFE fragment order randomization, referred to as MOSES-Augmented. Validation and test splits were consistent across all datasets. 

\subsubsection{Model}
We evaluated four autoregressive CLM architectures: an RNN as a baseline, GPT-2 similar to the original SAFE work, LLaMA \citep{touvron2023LLaMA}, and Jamba \citep{lieber2024Jamba}. GPT-2 uses absolute positional encoding, where fixed positional information is added to token embeddings. In contrast, LLaMA employs Rotary Positional Embeddings (RoPE), which rotates token embeddings based on their relative positions. RoPE improves the model's ability to capture long-range token dependencies by better preserving relationships across varying sequence lengths. Jamba is a novel hybrid Transformer-Mamba model claimed to increase model capacity while keeping active parameter usage low. After performing hyperparameter searches, we trained 92 models in total, across 4 architectures, 3-4 datasets, and 6 representations. Further details on training can be found in section \ref{appendix:arch-training} of the Appendix.

\subsection{Evaluation}
We assessed model performance in both \textit{de novo} generation and fragment-constrained tasks using standard metrics such as validity, uniqueness, and internal diversity \citep{polykovskiy2020molecular,Huang2021tdc,thomas2024molscore}, alongside SAFE-specific metrics like fragmentation percentage which measures the proportion of molecular graphs generated with disconnected fragments. For \textit{de novo} generation, we sampled 10,000 molecules across 5 seeds, reporting average performance. In fragment-constrained tasks, we evaluated scaffold decoration using the SureChEMBL (17 scaffolds) and DRD2 (5 scaffolds) benchmarks \citep{langevin2020scaffold}, and both scaffold decoration and fragment linking on the DRUG dataset (10 molecules) proposed in \citet{noutahi2024gotta}. The DRUG dataset serves as a challenging out-of-distribution benchmark due to its divergence from the training MOSES dataset. Any molecules or fragments with tokens not found in the respective vocabularies were excluded from evaluation. We benchmarked SAFE-based models against the closest SMILES-based frameworks, SAMOA and PromptSMILES. For fragment-constrained tasks, benchmarking was conducted using 5,000 samples per constraint, with all methods trained on MOSES-Full.

\section{Results}

\subsection{Effect of Model Architecture}

\begin{table}[ht!]
\caption{Performance comparison of SAFE and SMILES models on MOSES-Full (10k samples, 5 replicates). The best performance within each representation is highlighted in gray, with the overall best performance in red. \textbf{*} denotes results not statistically different from the best ($\alpha = 0.01$).}
\centering
\resizebox{1.0\linewidth}{!}{
\begin{tabular}{llcccccc}
\toprule
{Representation} & {Model} & $\uparrow$ Validity (\%) & $\uparrow$ Uniqueness (\%) & $\uparrow$ Novelty (\%) & $\uparrow$ Int.Div (\%) & $\downarrow$ Fragmented (\%) \\
\midrule
\multirow{4}{*}{SMILES} & Jamba & 0.995 ± 0.001 & 0.997 ± 0.000 & \cellcolor{lightgray!50}\textcolor{red}{1.000 ± 0.000*} & 0.838 ± 0.002 & \cellcolor{lightgray!50}\textcolor{red}{0.000 ± 0.000*} \\
 & LLaMA & \cellcolor{lightgray!50}{0.998 ± 0.001*} & 0.998 ± 0.001* & \cellcolor{lightgray!50}\textcolor{red}{1.000 ± 0.000*} & \cellcolor{lightgray!50}{0.842 ± 0.001} & \cellcolor{lightgray!50}\textcolor{red}{0.000 ± 0.000*} \\
 & GPT-2 & 0.994 ± 0.001 & 0.997 ± 0.000 & \cellcolor{lightgray!50}\textcolor{red}{1.000 ± 0.000*} & 0.834 ± 0.001 & \cellcolor{lightgray!50}\textcolor{red}{0.000 ± 0.000*} \\
 & RNN & 0.987 ± 0.001 & \cellcolor{lightgray!50}{0.999 ± 0.000*} & \cellcolor{lightgray!50}\textcolor{red}{1.000 ± 0.000*} & 0.835 ± 0.001 & \cellcolor{lightgray!50}\textcolor{red}{0.000 ± 0.000*} \\
\midrule
\multirow{4}{*}{SAFE-BRICS} & Jamba & 0.967 ± 0.001 & 0.998 ± 0.000 & 0.806 ± 0.005 & 0.847 ± 0.000 & 0.033 ± 0.002 \\
 & LLaMA & \cellcolor{lightgray!50}{0.990 ± 0.001} & 0.997 ± 0.000 & 0.765 ± 0.004 & 0.847 ± 0.000 & \cellcolor{lightgray!50}{0.014 ± 0.001} \\
 & GPT-2 & 0.970 ± 0.002 & 0.998 ± 0.000 & 0.858 ± 0.003 & 0.847 ± 0.000 & 0.029 ± 0.001 \\
 & RNN & 0.938 ± 0.003 & \cellcolor{lightgray!50}\textcolor{red}{1.000 ± 0.000*} & \cellcolor{lightgray!50}{0.915 ± 0.003} & \cellcolor{lightgray!50}{0.859 ± 0.000} & 0.105 ± 0.003 \\
\midrule
\multirow{4}{*}{SAFE-RECAP} & Jamba & 0.952 ± 0.004 & 0.998 ± 0.001 & 0.783 ± 0.006 & 0.847 ± 0.000 & 0.016 ± 0.002 \\
 & LLaMA & \cellcolor{lightgray!50}{0.991 ± 0.001} & 0.997 ± 0.000 & 0.739 ± 0.005 & 0.848 ± 0.000 & \cellcolor{lightgray!50}{0.005 ± 0.001} \\
 & GPT-2 & 0.978 ± 0.002 & 0.997 ± 0.001 & 0.803 ± 0.004 & 0.847 ± 0.001 & 0.006 ± 0.001 \\
 & RNN & 0.883 ± 0.002 & \cellcolor{lightgray!50}\textcolor{red}{1.000 ± 0.000*} & \cellcolor{lightgray!50}{0.878 ± 0.001} & \cellcolor{lightgray!50}{0.857 ± 0.000} & 0.039 ± 0.004 \\
\midrule
\multirow{4}{*}{SAFE-HR} & Jamba & 0.942 ± 0.002 & 0.999 ± 0.000 & 0.907 ± 0.002 & 0.852 ± 0.000 & 0.124 ± 0.004 \\
 & LLaMA & \cellcolor{lightgray!50}{0.974 ± 0.001} & 0.998 ± 0.000 & 0.868 ± 0.004 & 0.850 ± 0.001 & \cellcolor{lightgray!50}{0.064 ± 0.002} \\
 & GPT-2 & 0.925 ± 0.003 & 0.998 ± 0.000 & 0.902 ± 0.003 & 0.851 ± 0.000 & 0.088 ± 0.003 \\
 & RNN & 0.854 ± 0.002 & \cellcolor{lightgray!50}\textcolor{red}{1.000 ± 0.000*} & \cellcolor{lightgray!50}{0.989 ± 0.001} & \cellcolor{lightgray!50}\textcolor{red}{0.866 ± 0.000*} & 0.333 ± 0.005 \\
\midrule
\multirow{4}{*}{SAFE-MMPA} & Jamba & 0.967 ± 0.002 & 0.998 ± 0.000* & 0.888 ± 0.002 & 0.849 ± 0.000 & 0.095 ± 0.003 \\
 & LLaMA & 0.984 ± 0.001 & 0.998 ± 0.000 & 0.844 ± 0.002 & 0.848 ± 0.001 & \cellcolor{lightgray!50}{0.049 ± 0.002} \\
 & GPT-2 & 0.968 ± 0.002 & 0.998 ± 0.000 & 0.890 ± 0.003 & 0.849 ± 0.000 & 0.063 ± 0.002 \\
 & RNN & \cellcolor{lightgray!50}\textcolor{red}{1.000 ± 0.000*} & \cellcolor{lightgray!50}\textcolor{red}{1.000 ± 0.000*} & \cellcolor{lightgray!50}{0.967 ± 0.001} & \cellcolor{lightgray!50}{0.862 ± 0.000} & 0.250 ± 0.007 \\
\midrule

\multirow{4}{*}{SAFE-ROTATABLE} & Jamba & 0.959 ± 0.002 & 0.998 ± 0.000 & 0.877 ± 0.003 & 0.850 ± 0.000 & 0.064 ± 0.002 \\
 & LLaMA & \cellcolor{lightgray!50}{0.988 ± 0.001} & 0.997 ± 0.001 & 0.813 ± 0.003 & 0.849 ± 0.001 & \cellcolor{lightgray!50}{0.020 ± 0.001} \\
 & GPT-2 & 0.974 ± 0.002 & 0.998 ± 0.000 & 0.884 ± 0.003 & 0.849 ± 0.000 & 0.041 ± 0.002 \\
 & RNN & 0.919 ± 0.002 & \cellcolor{lightgray!50}\textcolor{red}{1.000 ± 0.000*} & \cellcolor{lightgray!50}{0.954 ± 0.002} & \cellcolor{lightgray!50}{0.861 ± 0.000} & 0.141 ± 0.002 \\
\bottomrule
\end{tabular}
}
\label{tab:1}
\end{table}

\begin{figure}[!ht]
    \centering
    \includegraphics[width=\linewidth]{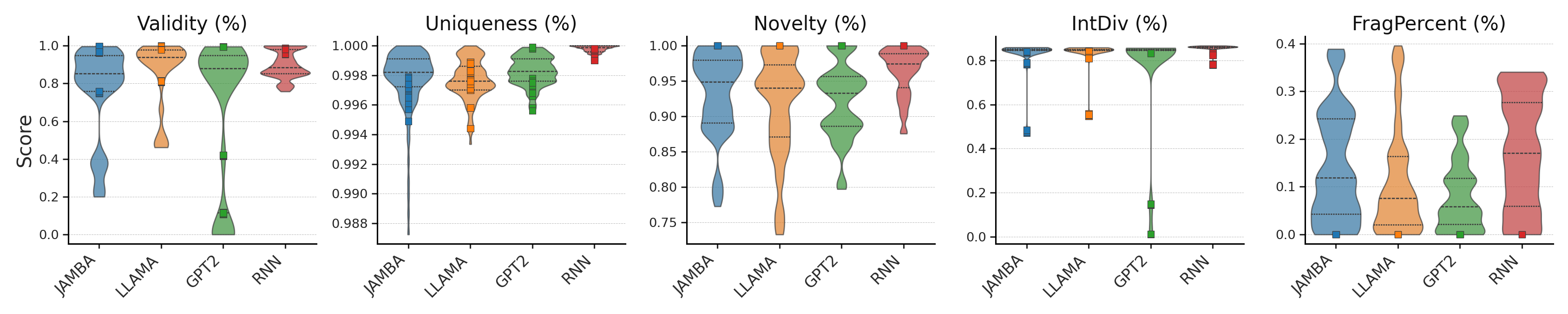}
     \caption{Performance of each architecture across 4 datasets and 6 representations. SMILES-based results (squares) are indicated for reference.}
    \label{fig:model-effect}
\end{figure}

Table~\ref{tab:1} compares the generative performance of the four architectures across the MOSES-Full dataset for each representation, and Figure~\ref{fig:model-effect} illustrates their overall performance across all datasets. Both show that while smaller RNN models are efficient at generating novel, diverse, and unique molecules, they exhibit limitations in validity and struggle with the complexities of the SAFE representation grammar. By contrast, LLaMA consistently generates the most valid and least fragmented molecules on SAFE strings, highlighting its superior capacity to model the underlying data distribution and grasp the syntax of the SAFE line notation. However on internal diversity, LLaMa models underperformed compared to RNNs. Intuitively, models that excel at fitting the data and can capture its intricate rules are also likely to experience a tradeoff, with a reduction in diversity and novelty. These observations align with previous findings that sampling generalization in CLMs (measured by internal diversity and novelty), does not necessarily correlate with validity~\citep{skinnider2024invalid}.

\subsection{Effect of Dataset Size and Data Augmentation}

\begin{figure}[!h]
    \centering
    \includegraphics[width=\linewidth]{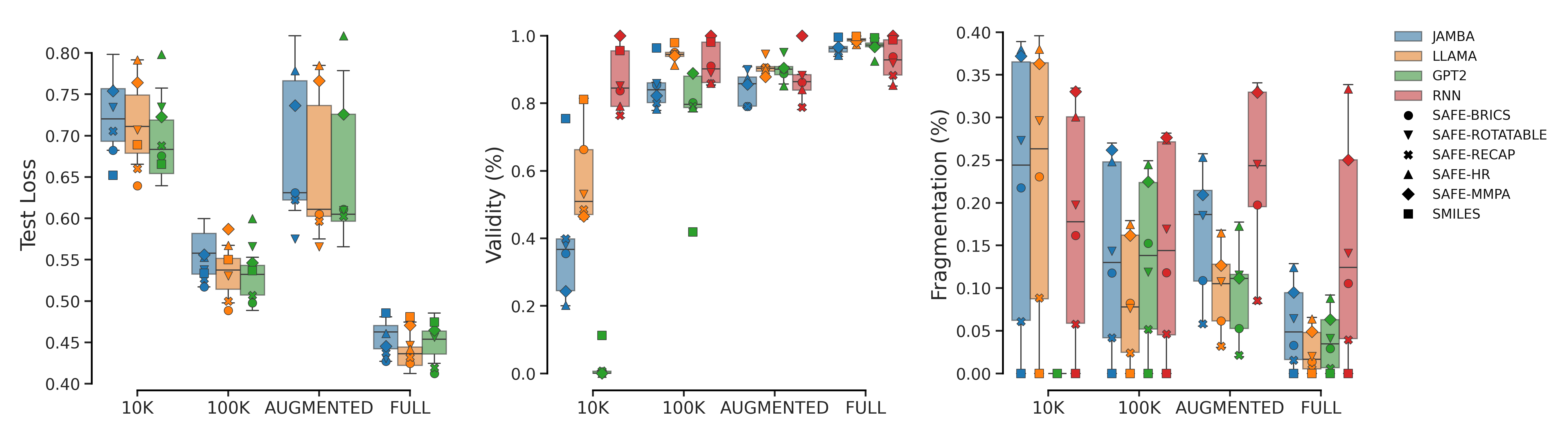}
    \caption{Performance across datasets of different sizes, measured by test loss, validity, and fragmentation percentage. RNNs' test loss results are omitted due to incomparable scale and the use of a different tokenization approach.}
    \label{fig:datasize-effect}
\end{figure}

Figure~\ref{fig:datasize-effect} shows that increasing dataset size generally reduces test loss and improves validity across all models and representations, indicating better generalization and modeling of the data distribution.  SAFE models also exhibit a decrease in fragmentation percentage with larger datasets, suggesting an improved understanding of the respective line notation grammar and cross-fragment linking schemes. Interestingly, RNN models show the least improvement  as datasets grow, struggling with SAFE syntax (evidenced by a high fragmentation percentage), despite maintaining high validity. In contrast, GPT-2-based models benefit significantly from increased data, with performance jumping from almost no valid molecules on MOSES-10k to 100\% validity on MOSES-Full, alongside gains in internal diversity.

We note, surprisingly, that unlike SAFE-based models, which can generate diverse molecules even in settings with limited training data, SMILES-based models collapse on internal diversity in low-data regimes. For instance, SMILES-GPT-2 shows poor diversity and validity, even on MOSES-100k, compared to their SAFE counterparts (Figure~\ref{fig:dataset-effect-per-model}).

While SAFE randomization helps maintain average validity and benefits data-hungry models like GPT-2, it does not significantly enhance generalization or improve comprehension of SAFE syntax. Although it is possible that further increasing the percentage of data augmentation could lead to improved performance, results observed on MOSES-Augmented suggest that the primary benefit of SAFE randomization augmentation may lie in extending training time and preventing overfitting. Expanding the amount of novel and diverse structures in the training set might yield similar results while also significantly improving the understanding of SAFE syntax.

Nevertheless, we note that data augmentation via SAFE randomization mitigates the usual trend of reduced novelty as training data size increases (Figure~\ref{fig:dataset-effect-per-model}). We hypothesize that, akin to SMILES randomization, SAFE randomization helps maintain novelty by exposing the model to alternative molecular configurations during training.

\subsection{Effect of Different Fragmentation Algorithms on Performance}

\begin{figure}[!hb]
    \centering
    \includegraphics[width=1\linewidth]{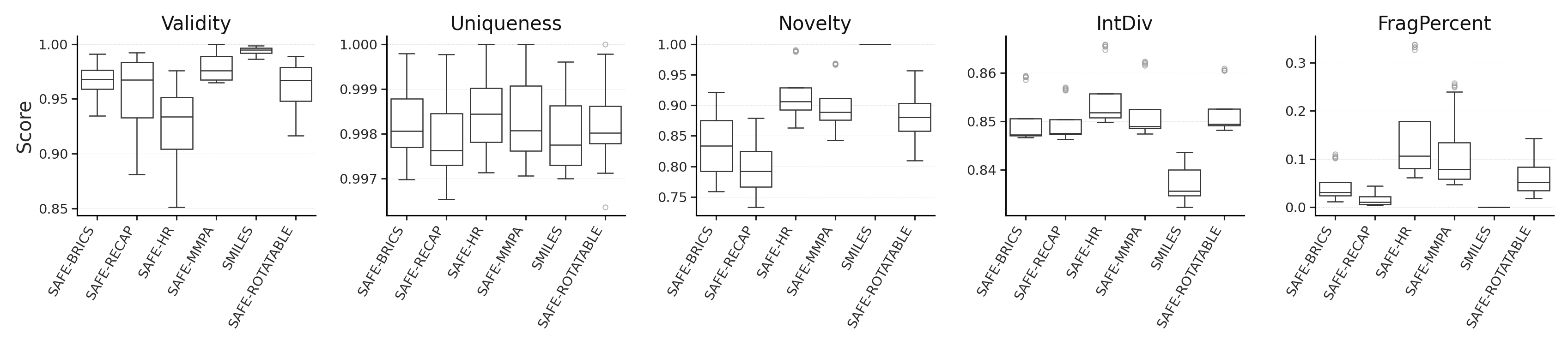}
    \caption{Effect of fragmentation algorithm on generative metrics and on the percentage of fragmented molecules, across all models, on MOSES-Full.}
    \label{fig:fragmentation-effect}
\end{figure}

Figure~\ref{fig:fragmentation-effect} compares the performance of SAFE generative models trained on MOSES-Full across various bond disconnection algorithms, evaluating how each affects their generative capabilities. Indeed, the SAFE grammar can introduce additional challenges compared to SMILES, as models must learn the cross-fragment linking schemes embedded in the grammar.

The results indicate that validity, internal diversity, novelty, and uniqueness are generally high across all bond disconnection methods. However, HR and RECAP models slightly underperform in validity.

\begin{figure}[!htp]
    \centering
    \includegraphics[width=1\linewidth]{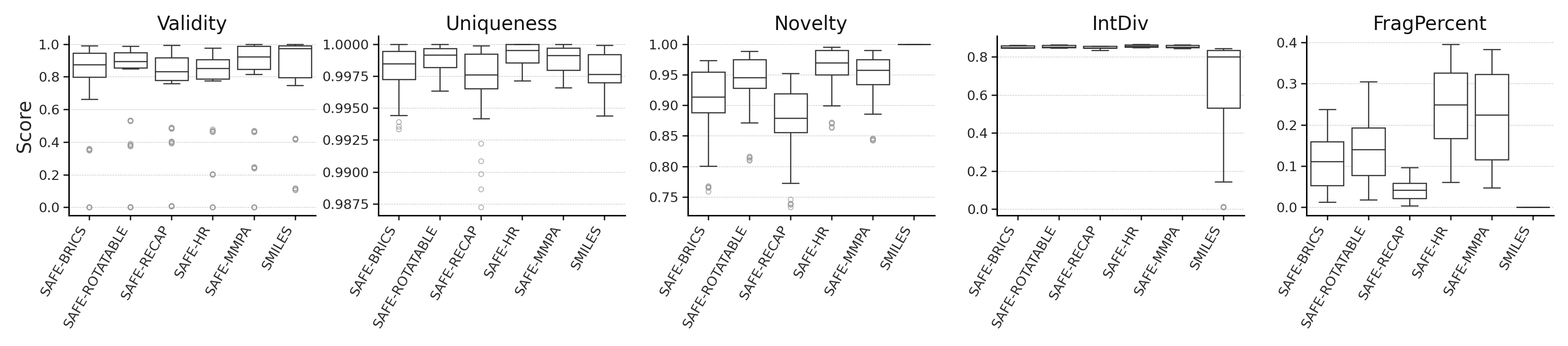}
    \caption{Effect of fragmentation algorithm on performance, across all models, when taking into account all datasets irrespective of the size.}
    \label{fig:fragmentation-effect-full}
\end{figure}

Though all fragmentation algorithms produce diverse and unique molecules, novelty scores are slightly lower than those of SMILES-based models This is likely due to SAFE’s more complex grammar and the need to learn specific tokens positions (e.g., '.' and digits) and their associations to maintain validity, which can, in turn impact generalization.  Notably, BRICS and RECAP decompositions result in significantly fewer fragmented molecules compared to HR and MMPA. This is however not because they exploit synthetic accessibility rules but rather, as shown in Figure~\ref{fig:frag-distribution}, because BRICS and RECAP bond disconnections produce fewer and larger fragments, leading to simpler grammar for models to learn. Thus, SAFE-BRICS and SAFE-RECAP represent syntaxes that are significantly easier to learn compared to other decompositions, as they require fewer cross-fragment linking tokens, making them excellent compromises for training SAFE-based CLMs. Additionally, on the simple MOSES-Full dataset, these decompositions yielded molecules with the highest QED and SAScore, comparable to or better than those generated by SMILES-based models (see Table~\ref{tab:2}). These observations are consistent across all datasets, as shown in Figure~\ref{fig:fragmentation-effect-full}.

\subsection{Performance on Fragment-Constrained Molecule Design Tasks}

\subsubsection{SAFE Outperforms SMILES on Scaffold Decoration}

\begin{figure}[!h]
    \centering
    \includegraphics[width=1\linewidth]{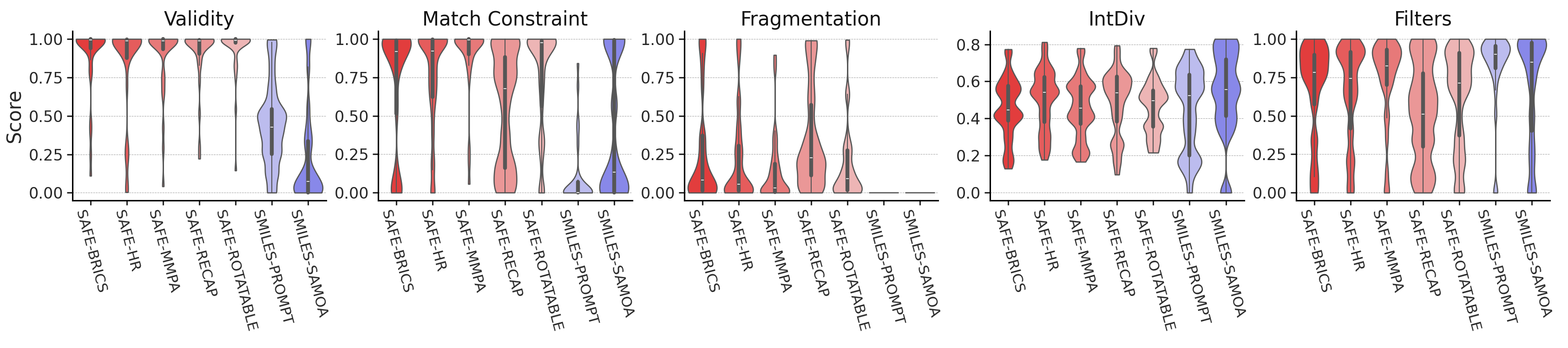}
    \caption{Average performance of SAFE and SMILES sampling algorithms on scaffold decoration tasks (3 benchmarks, 29 scaffolds). SAFE models outperformed SMILES approaches. SMILES-PROMPT refers PromptSMILES, and SMILES-SAMOA to SAMOA.}
    \label{fig:scaffold-dec-mean}
\end{figure}

Using LLaMA, the most stable and robust CLM architecture, we evaluated the performance of various SAFE-based models and two alternative SMILES approaches on standard scaffold decoration benchmarks. We assessed the generated molecules on several criteria: validity, preservation of the input scaffold constraint (\texttt{Match Constraint}), fragmentation percentage, internal diversity, and the proportion of molecules passing the medicinal chemistry filters used to curate the MOSES dataset. The latter metric helps gauge how well each algorithm captures implicit rules from the training data distribution when generating fragment-constrained samples. Figure \ref{fig:scaffold-dec-mean} presents the average performance of all scaffold decoration approaches evaluated.

SAFE models consistently outperformed SMILES-based approaches across all metrics. Despite generating fragmented molecules in some cases, SAFE-based methods produced the most valid molecules while preserving the input scaffold, unlike both PromptSMILES and SAMOA. Surprisingly, SAMOA generated the highest number of invalid molecules. Although around 50\% of the molecules generated by PromptSMILES were valid, they often failed to respect scaffold constraints. Both SMILES-based algorithms performed particularly poorly on the DRUG dataset (see Figure~\ref{fig:dataset-effect-per-model}) outright failing on several scaffolds (see Figure~\ref{fig:per-benchmark-scaffold-dec}), suggesting they struggle to generalize to out-of-distribution scaffolds or chemical structures.

Upon closer inspection of molecules generated for the drug Baricitinib under standardized scaffold constraints: \textbf{\texttt{[*]N1CC([*])(n2cc(-c3ncnc4[nH]ccc34)cn2)C1}} (see Figure~\ref{fig:scaffold-dec-baricitinib}), we observed that SAMOA's sampling algorithm rewrote the scaffold by forming new bonds between existing atoms. Due to early stopping during autoregressive generation (either from sampling the \texttt{<EOS>} token or reaching the maximum sequence length), SAMOA can produce random and divergent scaffolds lacking significant parts of the input scaffold. While PromptSMILES, which rearranges the scaffold to position the attachment point last, was more robust, it still generated different, though structurally closer, scaffolds.

Among SAFE-based models, RECAP showed the lowest performance on fragmentation percentage and scaffold preservation metrics, additionally struggling to maintain the implicit filtering rules of the training data. Other fragmentation algorithms, especially MMPA, were highly consistent in both standard generative metrics and molecular quality (as measured by SAScore and QED) (see Figure~\ref{fig:per-benchmark-scaffold-dec-alt}). SAFE-HR, meanwhile, produced the most diverse decorations overall.

\subsubsection{SAFE Outperforms SMILES on Linker Design}

\begin{figure}[!h]
    \centering
    \includegraphics[width=1\linewidth]{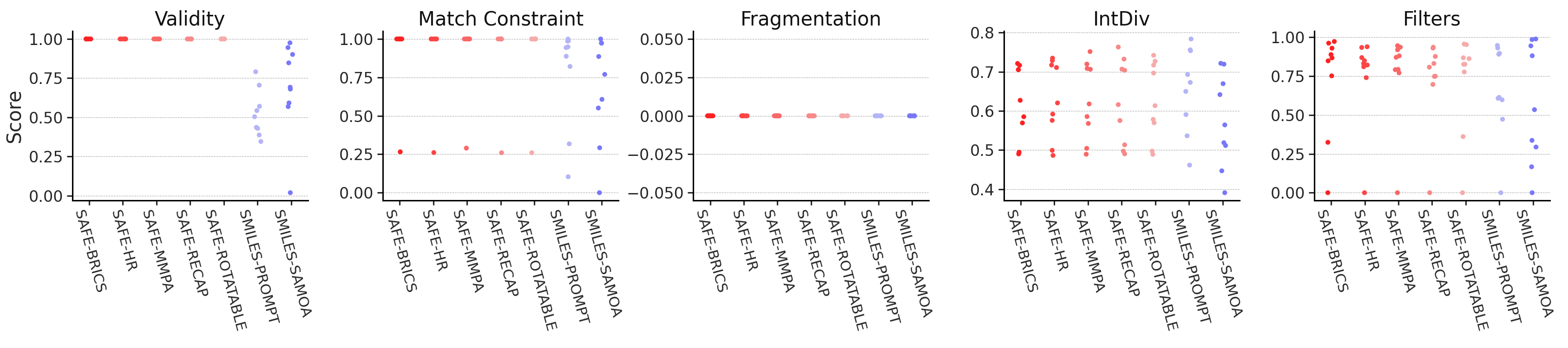}
    \caption{Average performance of SAFE and SMILES algorithms on linker design tasks (9 fragment sets from the DRUG benchmark).}
    \label{fig:linker-design-mean}
\end{figure}

As in the scaffold decoration task, Figure~\ref{fig:linker-design-mean} and show that SAFE-based methods outperformed SMILES-based approaches in linker design. SAFE models achieved perfect validity and fragmentation scores, near-perfect substructure constraints preservation, and maintained high QED and SAScore (Figure \ref{fig:linker-design-alt}). In contrast, SMILES-based methods often collapsed, failing to maintain the input fragment constraints, as illustrated in Figure~\ref{fig:linker-design-cyclothiazide}~and~\ref{fig:per-linker-design}. SAFE approaches preserved these constraints, though occasionally sacrificing design diversity, especially on the most challenging inputs.

We hypothesize that the perfect validity of SAFE models can be can be partly attributed to LLaMA’s ability to better capture the underlying line notation grammar. This allows for 0-size linking by directly generating ring closures between the two unlinked fragments on difficult samples. Notably, the SAMOA algorithm also exploits this bias, as seen in the linker design examples for Baricitinib (Figure~\ref{fig:linker-design-baricitinib}).

\section{Discussion}

Our exploration of SAFE-based generative models demonstrated clear advantages over SMILES-based approaches, specifically on fragment-constrained molecule design tasks. While both approaches performed comparably in pure \textit{de novo} generation, SAFE-based models significantly outperformed SMILES-based methods in scaffold decoration and linker design.

SAFE-LLaMA models, in particular, demonstrated more robust performance across various representations, dataset sizes, and evaluation tasks. We attribute this, in part, to LLaMA’s use of Rotary Positional Embedding (RoPE), which likely captures positional dependencies between tokens of different fragments more effectively than GPT-2's absolute positional embeddings. This allowed LLaMA to better learn the nuances of SAFE grammar, resulting in the generation of more valid and connected molecular graphs. We did not find any specific advantage of Jamba over other architectures, but it demonstrated greater robustness on smaller datasets compared to GPT-2. This result likely reflects Jamba's design, which balances model capacity with parameter efficiency, making it less prone to overfitting in small datasets. Further investigation into its performance on datasets with larger, more complex molecules (e.g., polymers) could reveal potential strengths in those contexts.

We found that dataset size played a crucial role, with larger datasets consistently improving performance, especially for transformer models, and most notably for GPT-2. This finding emphasizes the importance of diverse and extensive training data for capturing a broader chemical space, to enable models to generalize more effectively in both \textit{de novo} and fragment-constrained tasks. While SAFE-based data augmentation did not significantly enhance generalization or sampling quality, it did help mitigate the loss of novelty in generated samples as dataset size increased, primarily by preventing overfitting. Investigating the potential synergy between SMILES and SAFE randomization may provide additional insights in future work.

Finally, the choice of fragmentation algorithm also had a significant impact on performance. SAFE-BRICS and SAFE-RECAP, which generate fewer and larger fragments using synthetic accessibility decomposition rules, were easier for models to learn, leading to lower fragmentation rates and improved synthetic accessibility of generated molecules. However, this simplicity came at a cost: models trained on decompositions with more fragments, like HR and MMPA, demonstrated higher novelty and uniqueness although at the expense of lower validity and increased fragmentation in the generated molecules. Interestingly, SAFE-RECAP performed the worst in scaffold decoration tasks, producing the most fragmented and least diverse molecules  (see Figure~\ref{fig:per-benchmark-scaffold-dec}). This may be due to RECAP’s rigid bond disconnection rules, limiting bond formation possibilities in fragment-constrained sampling when trained on a dataset like MOSES. By contrast, BRICS which expands the bond disconnection criteria used by RECAP from 11 to 16 by taking into account additional chemical environment surrounding bonds, has led to improved scores. Overall, our results suggest a balance between fragmentation complexity and generalization: fewer fragments simplify grammar but limit exploration of chemical space, especially in fragment-constrained design tasks. The performance of SAFE-ROTATABLE, which serves as a middle ground, supported this trade-off. Nevertheless, considering auxiliary objectives like synthetic accessibility and drug-likeness, we recommend SAFE-BRICS as the most effective representation.

\section{Conclusion and Future Works}

While this study focused on finding the best training setup for SAFE generative models, several promising avenues for future research are apparent. One interesting direction is ensuring that the distribution of randomized SAFE strings preserves a consistent probabilistic density within the generative model’s learned distribution. This could enhance the model’s ability to effectively capture the underlying chemical space. Another important challenge is improving stereochemistry handling, which remains an issue for both SMILES and SAFE representations. Addressing these challenges could pave the way for using SAFE strings in representation learning, resulting in new opportunities for predictive and unsupervised tasks in molecular design.

Additionally, a deeper exploration of the robustness of SAFE-based models in optimization, particularly in low-data settings where diversity and novelty are crucial, is needed. While prior work used Proximal Policy Optimization (PPO) for goal-directed optimization of a SAFE-GPT model, other optimization algorithms should be systematically evaluated and benchmarked. Furthermore, while this study did not examine the effects of scaling model parameters, leveraging the novel insights gained here, it would be worthwhile to explore scaling SAFE models trained on large, diverse datasets spanning a wide range of molecular structures, properties, and sizes. This could lead to foundational models with significant utility for generative molecular design.

In conclusion, SAFE-based generative CLMs offer a stable, versatile, and powerful alternative to SMILES for molecular design, with considerable potential to advance applications in drug discovery and material design. We believe future research should focus on experimentally validating these models in practical settings to assess their effectiveness and real-world impact.

\bibliography{valence_bib}
\bibliographystyle{valence}
\clearpage
\appendix
\section{Appendix}

This appendix provides additional details regarding the dataset, model architectures, training configurations, and experiment results, supplementing the findings discussed in the main paper.

\subsection{Dataset}
\begin{figure}[!h]
    \centering
    \includegraphics[width=0.85\linewidth]{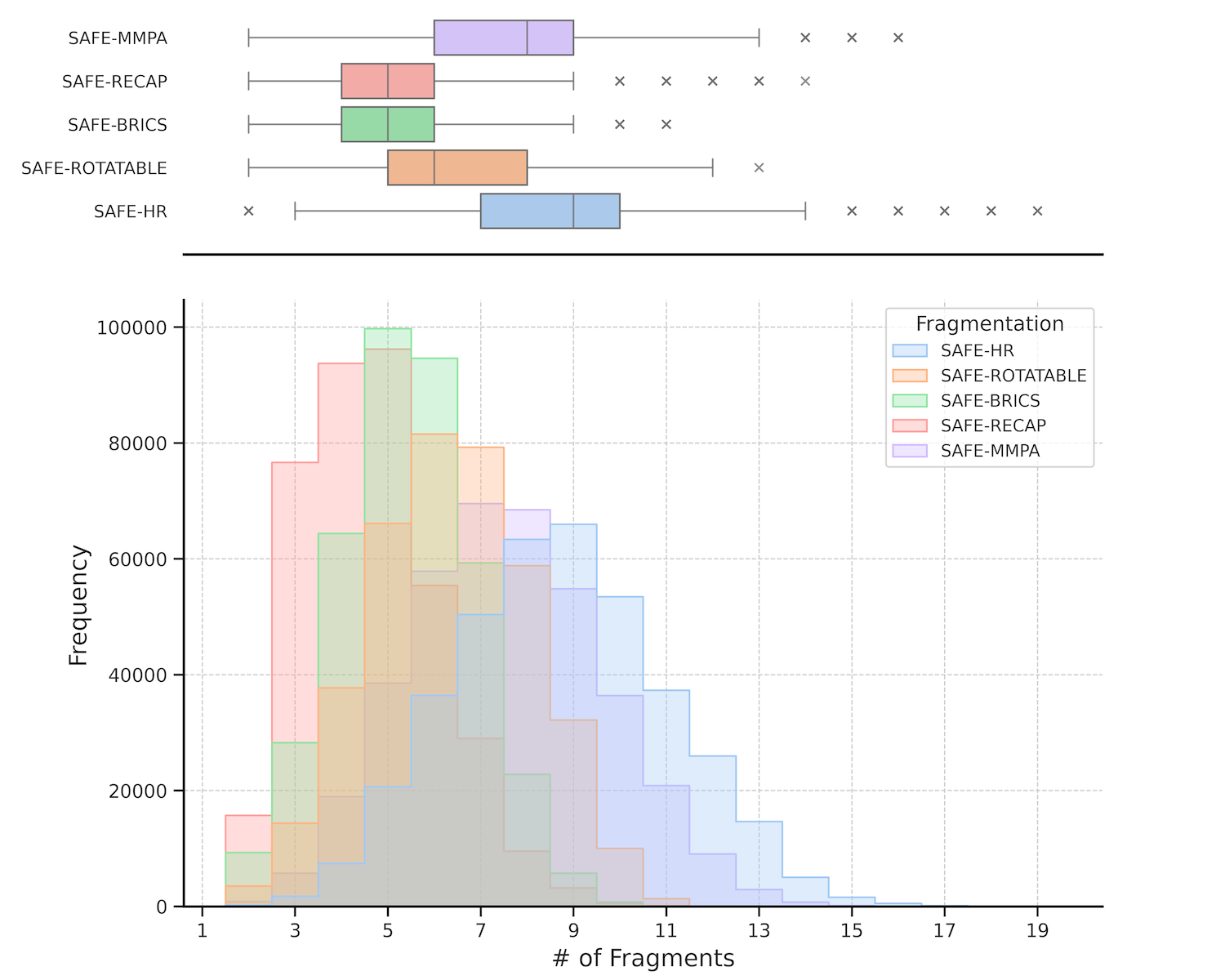}
    \caption{Distribution of the number of fragments for each bond disconnection algorithm.}
    \label{fig:frag-distribution}
\end{figure}

Figure~\ref{fig:frag-distribution} shows the distribution of the number of fragments generated by the different bond disconnection algorithms used to create the SAFE versions of the MOSES dataset. Both RECAP and BRICS algorithms resulted in molecules with fewer fragments (average of 5 fragments), while the Hussain-Rea (HR) and MMPA decompositions produced the highest number of fragments on average (9 and 8 fragments, respectively). The distribution of the number of fragments generated by these algorithms plays an essential role in the complexity of learning SAFE representations.

\subsection{Model Architecture and Training}
\label{appendix:arch-training}

\begin{figure}[!h]
    \centering
    \includegraphics[width=1\linewidth]{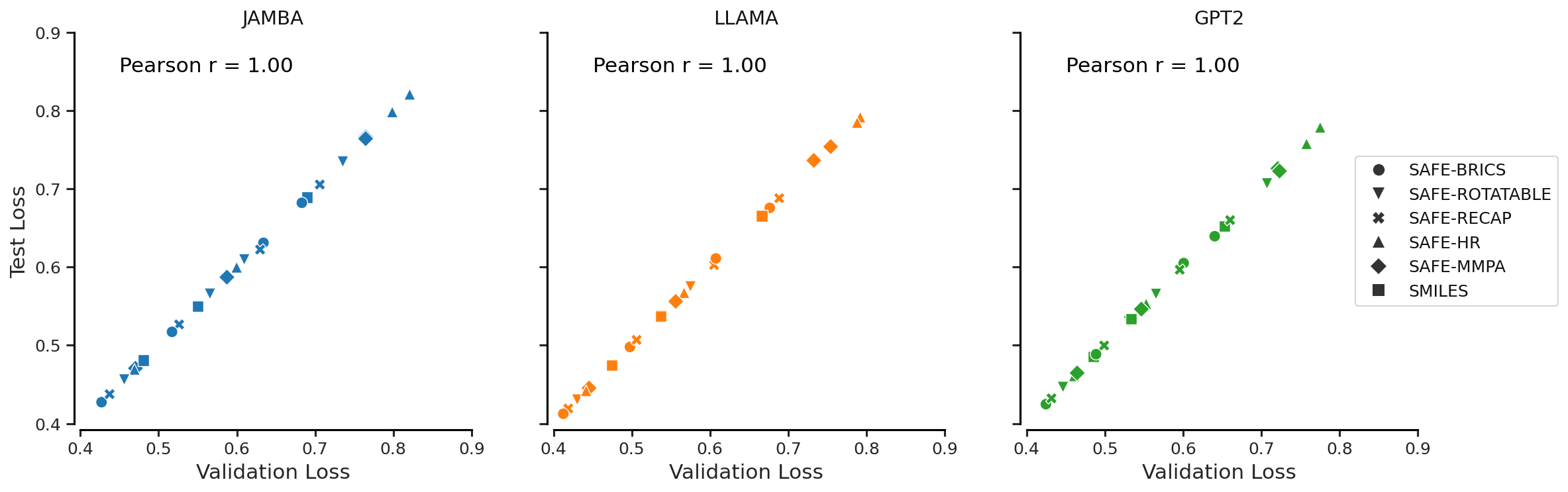}
    \caption{Validation vs Test Loss Across Different Transformer Architectures}
    \label{fig:app-valid-vs-test-loss}
\end{figure}

For each architecture, we performed a grid search on key hyperparameters such as learning rate, batch size, gradient norm clipping, and model configurations to ensure comparable sizes across all transformer models. We used the codebase provided by \citet{noutahi2024gotta}~\footnote{\url{https://github.com/datamol-io/safe/}} for training the transformer models, and for RNN (GRU) models, we followed the setup proposed by \citet{thomas2024promptsmiles} using the SMILES-RNN codebase~\footnote{\url{https://github.com/MorganCThomas/SMILES-RNN/}}.

The models were trained on the various MOSES datasets, and for all transformer models, we used a batch size of 128 and trained them for up to 100,000 steps (5,000 steps on MOSES-10k). 
Optimization was performed using the AdamW optimizer with a cosine learning-rate scheduler and a 10\% warmup. RNN models were trained using an exponential learning-rate decay, with early stopping based on validation loss. We observed overfitting with large RNN models, so their size was kept smaller than that of the other models.  Each training run was executed on a single NVIDIA H100/80Gb GPU, utilizing approximately 20\% of its capacity. On average, model training took 10 hours. 

The token vocabulary was fixed for each representation/fragmentation algorithm, with size ranging from 37 (SMILES) to 53 (SAFE-BRICS) tokens for transformers, and from 26 (SMILES) to 38 (MMPA) for RNNs (the SMILES-RNN codebase use a different tokenization scheme).  

Below are the final configurations for each model:

\textbf{GPT-2 ($\sim$ 10M parameters):} 3 layers, 512 hidden size, 8 attention heads, 1024 max position embeddings, vocab size ranging from 37 to 53, GELU activation function, and 0.1 dropout.

\textbf{LLaMA ($\sim$ 9M parameters):} 4 hidden layers, 512 hidden size, 512 intermediate size, 8 attention heads, 1024 max position embeddings, 37 to 53 vocab size, SiLU activation function, 0.1 dropout, and RoPE theta of 500,000.0.

\textbf{Jamba ($\sim$ 9,4M parameters):} 2 layers, 480 hidden size, 512 intermediate size, 8 attention heads, 1024 max position embeddings, 37 to 53 vocab size, SiLU activation function, 0.1 dropout, with Mamba parameters (D-Conv: 4, D-State: 16).

\textbf{GRU ($\sim$ 4.3M parameters):} 3 layers, 512 hidden size, 256 embedding layer size, 0.0 dropout, trained with a batch size of 125, learning rate of 0.001, and 3-10 epochs depending on dataset size.

Figure \ref{fig:app-valid-vs-test-loss}  shows the relationship between validation loss and test loss across the three transformer architectures (Jamba, LLaMA, and GPT-2). Each marker represents a unique configuration using either SMILES or one of the SAFE fragmentations (BRICS, ROTATABLE, RECAP, HR, MMPA). The strong linear correlation (Pearson's r = 1.00) indicates that validation loss reliably predicts test loss, suggesting that the models generalize well from validation to test sets.

\subsection{Experiments}
\subsection{De novo generation}

We evaluated the performance of the generative models on \textit{de novo} molecule generation across the MOSES-Full dataset. Table \ref{tab:2} presents a comparison of key molecular properties, including QED, SAScore, cLogP, and Molecular Weight, across 50,000 sampled molecules (5 seeds, 10,000 samples per seed). All architectures and representations mostly performed similarly on these metrics. As expected, SMILES and SAFE representation using fragmentation algorithms that leverage synthetic accessibility rules (SAFE-BRICS, SAFE-RECAP) produced molecules that have marginally better SAScore in average.

\begin{table}[htb!]
\centering
\caption{Comparison of the sampled molecules (10k samples, 5 replicates) for each generative CLM performance trained on MOSES-Full dataset on QED, SAScore, cLogP, and Molecular Weight. The best performance within each representation is highlighted in gray, and the overall best performance across all representations is in red. All architecture and representation performed similarly.}
\centering
\resizebox{1.0\linewidth}{!}{
\begin{tabular}{llcccc}
\toprule
Representation & Model & QED  $\uparrow$ & SAScore  $\downarrow$ & cLogP & Mol. Weight \\
\midrule
\multirow{4}{*}{SMILES}        & Jamba & 0.816 ± 0.090 & 2.357 ± 0.449 & 2.505 ± 0.915 & 307.675 ± 27.445 \\
                               & LLaMA & 0.815 ± 0.090 & 2.368 ± 0.448 & 2.493 ± 0.908 & 308.132 ± 27.110 \\
                               & GPT-2  & \cellcolor{lightgray}{\textcolor{red}{0.819 ± 0.088}} & \cellcolor{lightgray}{\textcolor{red}{2.340 ± 0.446}} & 2.483 ± 0.914 & 306.920 ± 27.339 \\
                               & RNN   & 0.807 ± 0.096 & 2.451 ± 0.467 & 2.451 ± 0.959 & 306.517 ± 28.904 \\
\midrule
\multirow{4}{*}{SAFE-BRICS}    & Jamba & \cellcolor{lightgray}{0.815} ± 0.093 & 2.363 ± 0.494 & 2.572 ± 0.888 & 304.814 ± 27.848 \\
                               & LLaMA & 0.815 ± 0.092 & \cellcolor{lightgray}{2.357} ± 0.472 & 2.572 ± 0.881 & 305.547 ± 27.510 \\
                               & GPT-2  & 0.808 ± 0.097 & 2.386 ± 0.486 & 2.642 ± 0.915 & 311.903 ± 28.573 \\
                               & RNN   & 0.802 ± 0.097 & 2.485 ± 0.581 & 2.469 ± 0.988 & 307.134 ± 29.936 \\
\midrule
\multirow{4}{*}{SAFE-RECAP}    & Jamba & \cellcolor{lightgray}{0.818} ± 0.092 & 2.382 ± 0.484 & 2.545 ± 0.907 & 305.297 ± 28.945 \\
                               & LLaMA & 0.817 ± 0.092 & \cellcolor{lightgray}{2.379} ± 0.466 & 2.544 ± 0.889 & 306.980 ± 27.653 \\
                               & GPT-2  & 0.811 ± 0.095 & 2.380 ± 0.463 & 2.565 ± 0.907 & 312.147 ± 29.510 \\
                               & RNN   & 0.801 ± 0.100 & 2.508 ± 0.518 & 2.474 ± 0.977 & 307.473 ± 30.869 \\
\midrule
\multirow{4}{*}{SAFE-HR}       & Jamba & 0.801 ± 0.100 & 2.489 ± 0.612 & 2.610 ± 0.911 & 304.228 ± 28.523 \\
                               & LLaMA & \cellcolor{lightgray}{0.807} ± 0.096 & \cellcolor{lightgray}{2.406} ± 0.545 & 2.591 ± 0.907 & 304.863 ± 28.160 \\
                               & GPT-2  & 0.803 ± 0.099 & 2.410 ± 0.549 & 2.588 ± 0.907 & 307.660 ± 28.931 \\
                               & RNN   & 0.768 ± 0.116 & 2.980 ± 0.738 & 2.484 ± 0.986 & 307.896 ± 29.196 \\
\midrule
\multirow{4}{*}{SAFE-MMPA}     & Jamba & 0.809 ± 0.095 & 2.428 ± 0.583 & 2.617 ± 0.896 & 305.708 ± 28.510 \\
                               & LLaMA & \cellcolor{lightgray}{0.811} ± 0.093 & \cellcolor{lightgray}{2.373} ± 0.528 & 2.616 ± 0.898 & 305.985 ± 28.033 \\
                               & GPT-2  & 0.805 ± 0.097 & 2.387 ± 0.542 & 2.599 ± 0.907 & 310.175 ± 29.361 \\
                               & RNN   & 0.789 ± 0.106 & 2.794 ± 0.696 & 2.463 ± 0.977 & 308.580 ± 30.096 \\

\midrule
\multirow{4}{*}{SAFE-ROTATABLE} & Jamba & 0.812 ± 0.095 & 2.406 ± 0.537 & 2.639 ± 0.913 & 304.755 ± 28.853 \\
                               & LLaMA & \cellcolor{lightgray}{0.815} ± 0.093 & \cellcolor{lightgray}{2.362} ± 0.496 & 2.608 ± 0.903 & 305.078 ± 27.984 \\
                               & GPT-2  & 0.810 ± 0.096 & 2.404 ± 0.511 & 2.626 ± 0.927 & 310.809 ± 28.912 \\
                               & RNN   & 0.796 ± 0.102 & 2.690 ± 0.624 & 2.494 ± 0.994 & 307.297 ± 30.027 \\

\bottomrule
\end{tabular}
}
\label{tab:2}
\end{table}

\subsubsection{Impact of Data Augmentation}

\begin{figure}[!htp]
    \centering
    \includegraphics[width=\linewidth]{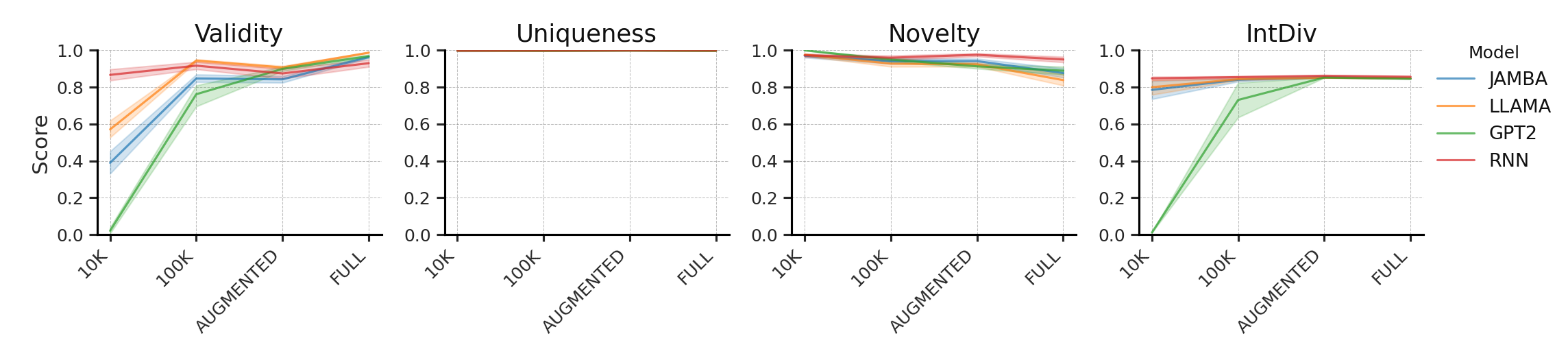}
    \caption{Effect of training data size on architecture performance, aggregated across representations.}
    \label{fig:mean-dataset-effect-per-model}
\end{figure}

\begin{figure}[!hb]
    \centering
    \includegraphics[width=\linewidth]{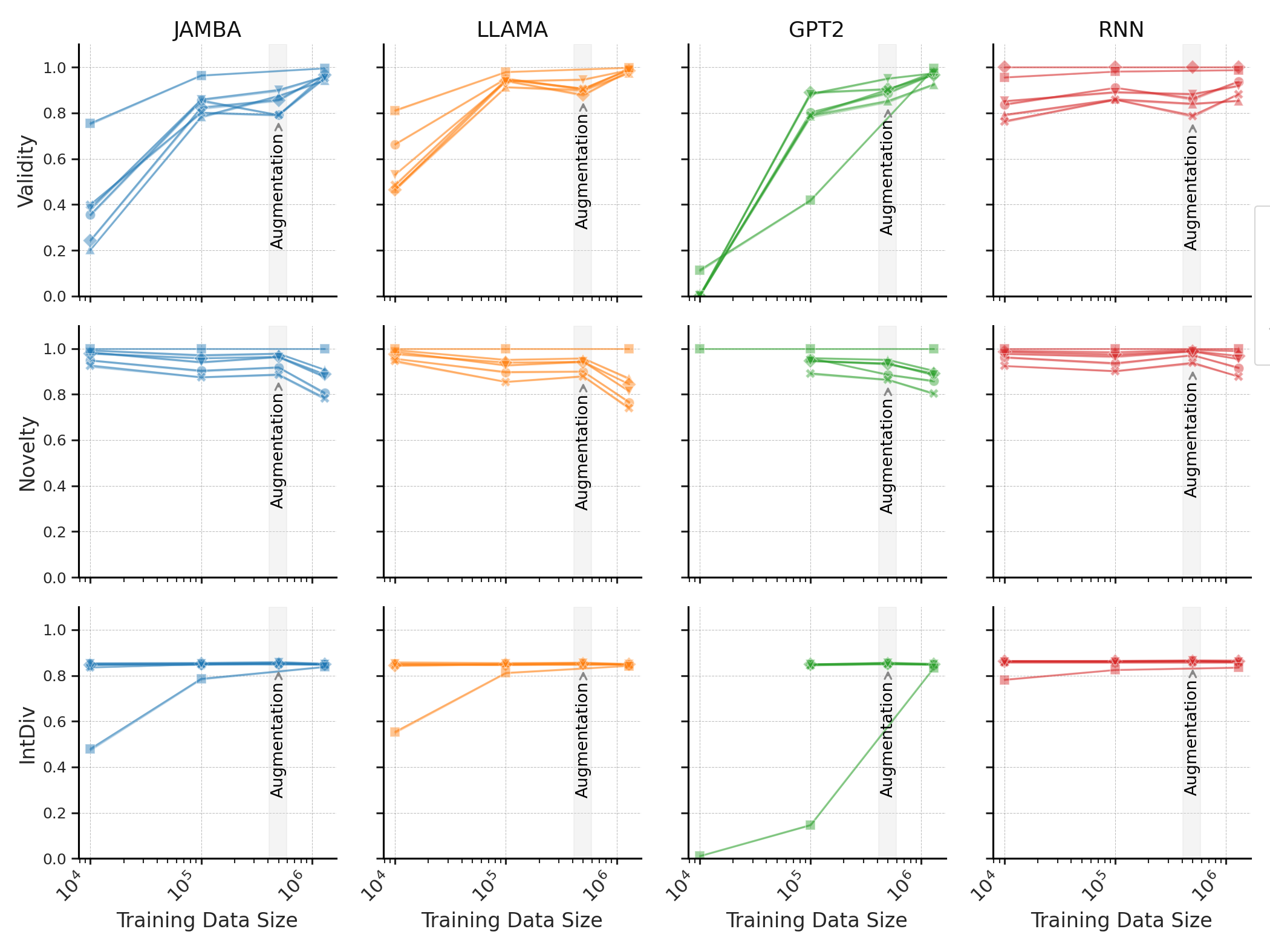}
    \caption{Performance of each architecture across different data sizes and representations.}
    \label{fig:dataset-effect-per-model}
\end{figure}

Figures \ref{fig:mean-dataset-effect-per-model} and \ref{fig:dataset-effect-per-model} illustrate the impact of dataset size and data augmentation on model performance. These results show that as dataset size increases, model performance usually improve. However, data augmentation does not provide equal benefits for all models. While augmentation helps maintain performance in low-data regimes by exposing models to diverse molecular configurations, its advantages vary depending on the model architecture.

\subsection{Fragment-Constrained Generation}

In this section, we evaluate the performance of various fragment-constrained generation methods. We focus on both scaffold decoration and linker design. The figures provide an in-depth look into how each model performs in preserving structural constraints while generating valid and chemically diverse molecules.

\begin{figure}[!ht]
    \centering
    \includegraphics[width=\linewidth]{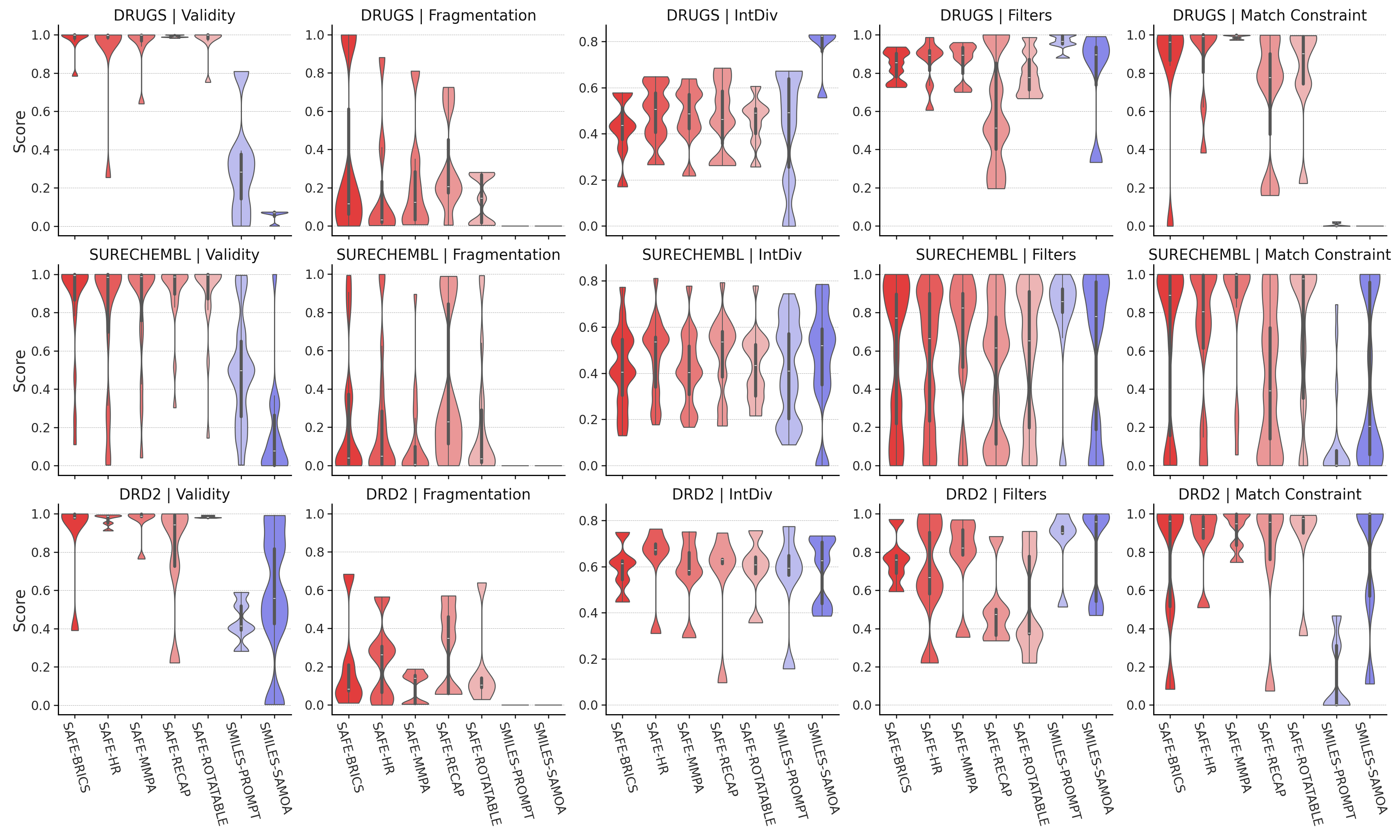}
    \caption{Per dataset performance of each fragment-constrained design method on scaffold decoration.}
    \label{fig:per-benchmark-scaffold-dec}
\end{figure}

\begin{figure}[!h]
    \centering
    \includegraphics[width=1\linewidth]{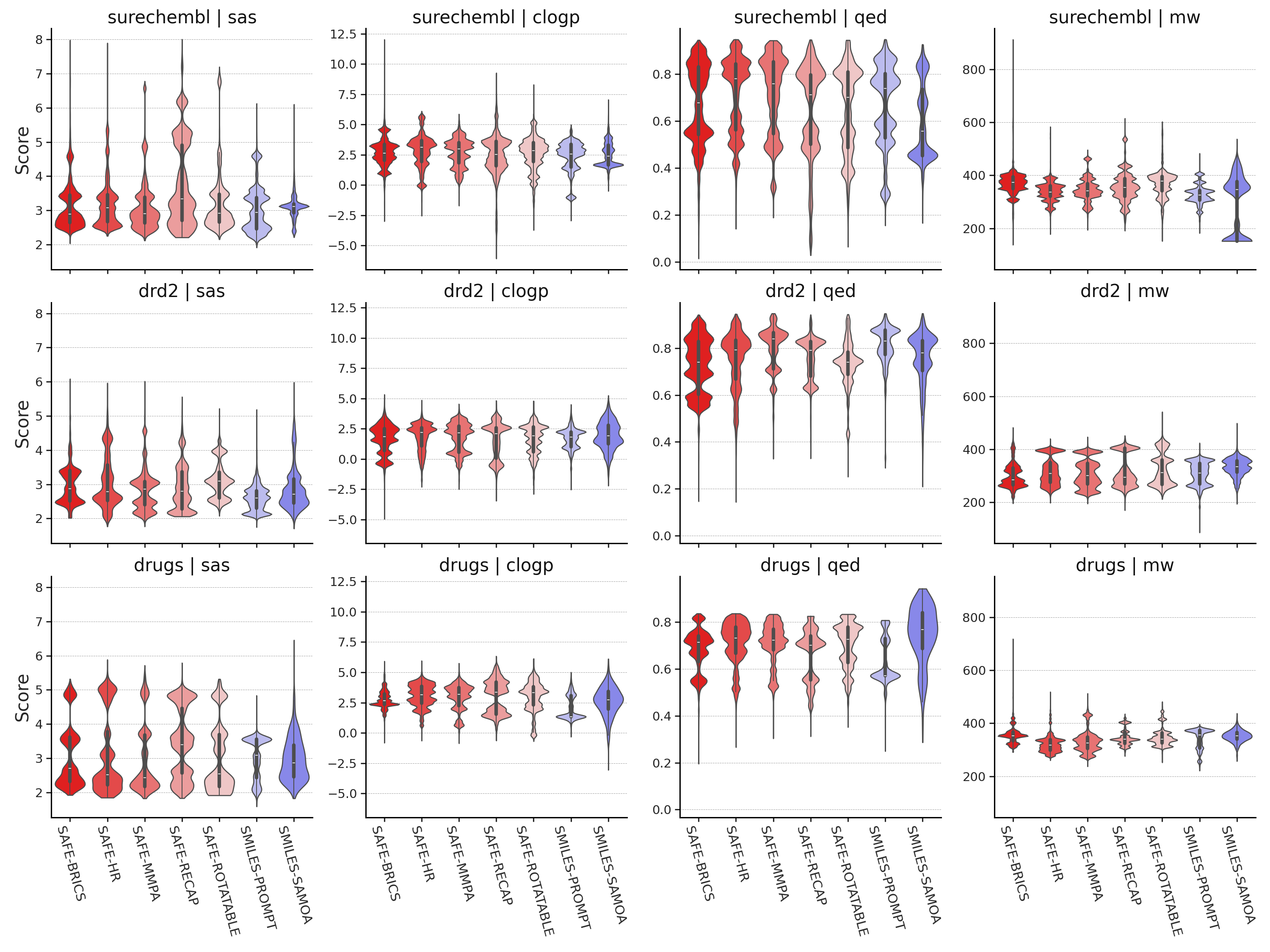}
    \caption{Per dataset standard molecule quality metrics distribution for each fragment-constrained design method on scaffold decoration.}
    \label{fig:per-benchmark-scaffold-dec-alt}
\end{figure}

\begin{figure}[!h]
    \centering
    \includegraphics[trim={0 0 4.5cm 0},clip, width=1\linewidth]{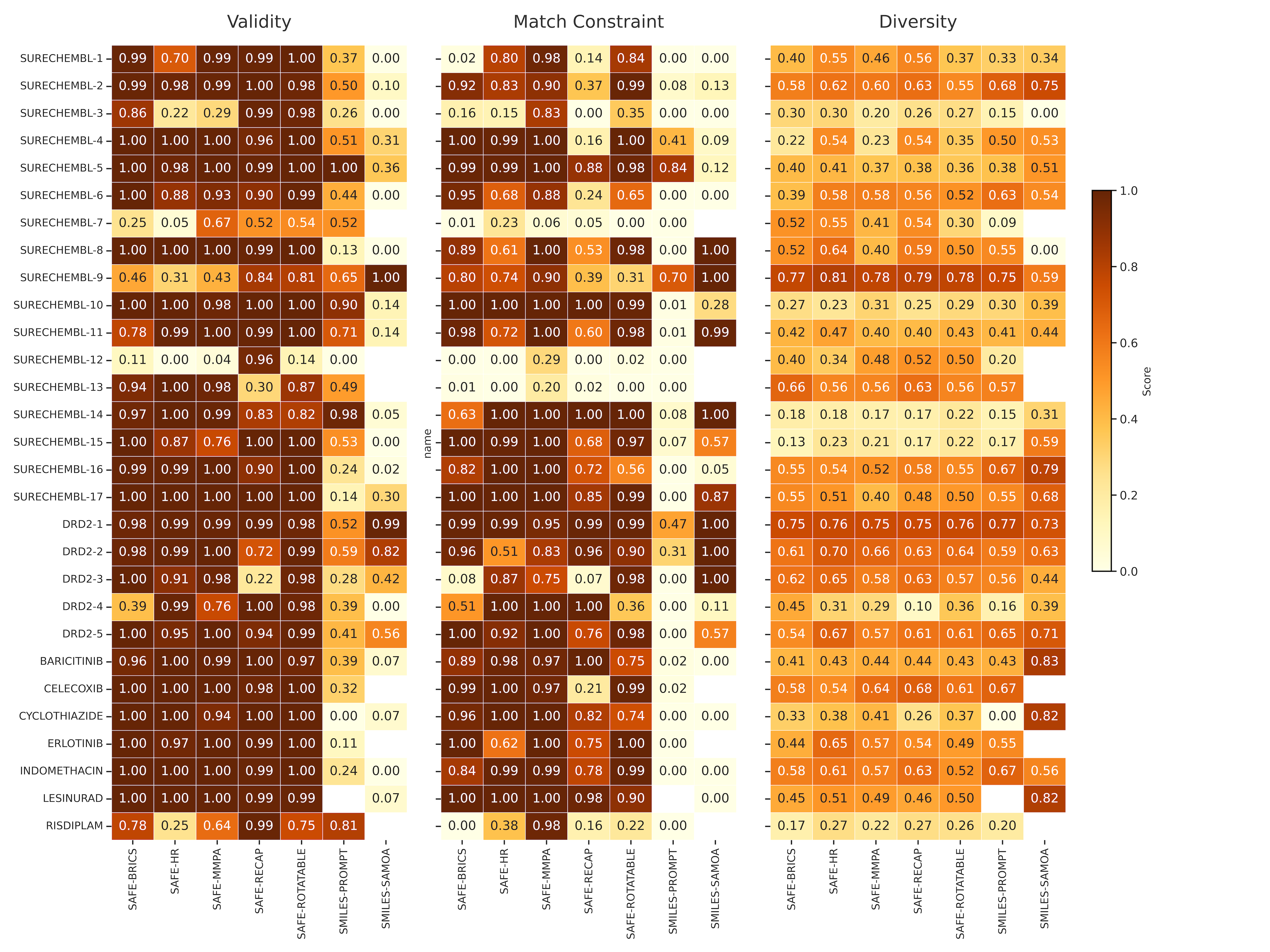}
    \caption{Per-scaffold performance of each fragment-constrained design method on the 3 different benchmarks considered for scaffold decoration.}
    \label{fig:per-scaffold-dec}
\end{figure}

\begin{figure}[!h]
    \centering
    \includegraphics[width=1\linewidth]{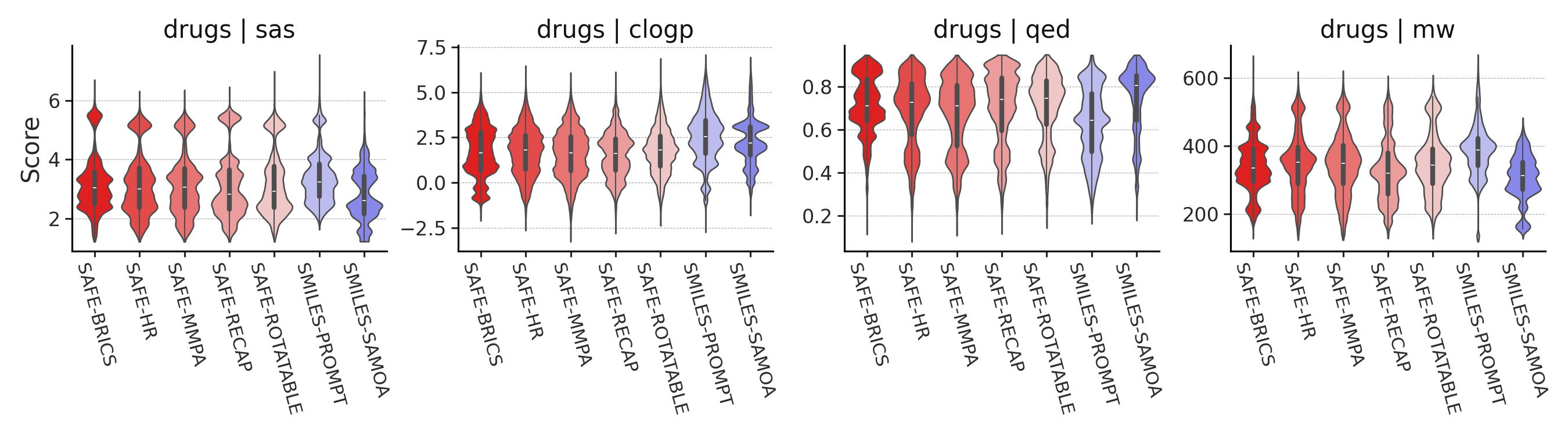}
    \caption{Performance on standard molecule quality metrics for each fragment-constrained design method on the DRUG linker design benchmark.}
    \label{fig:linker-design-alt}
\end{figure}

\begin{figure}[!h]
    \centering
    \includegraphics[trim={0 0 4.5cm 0},clip, width=1\linewidth]{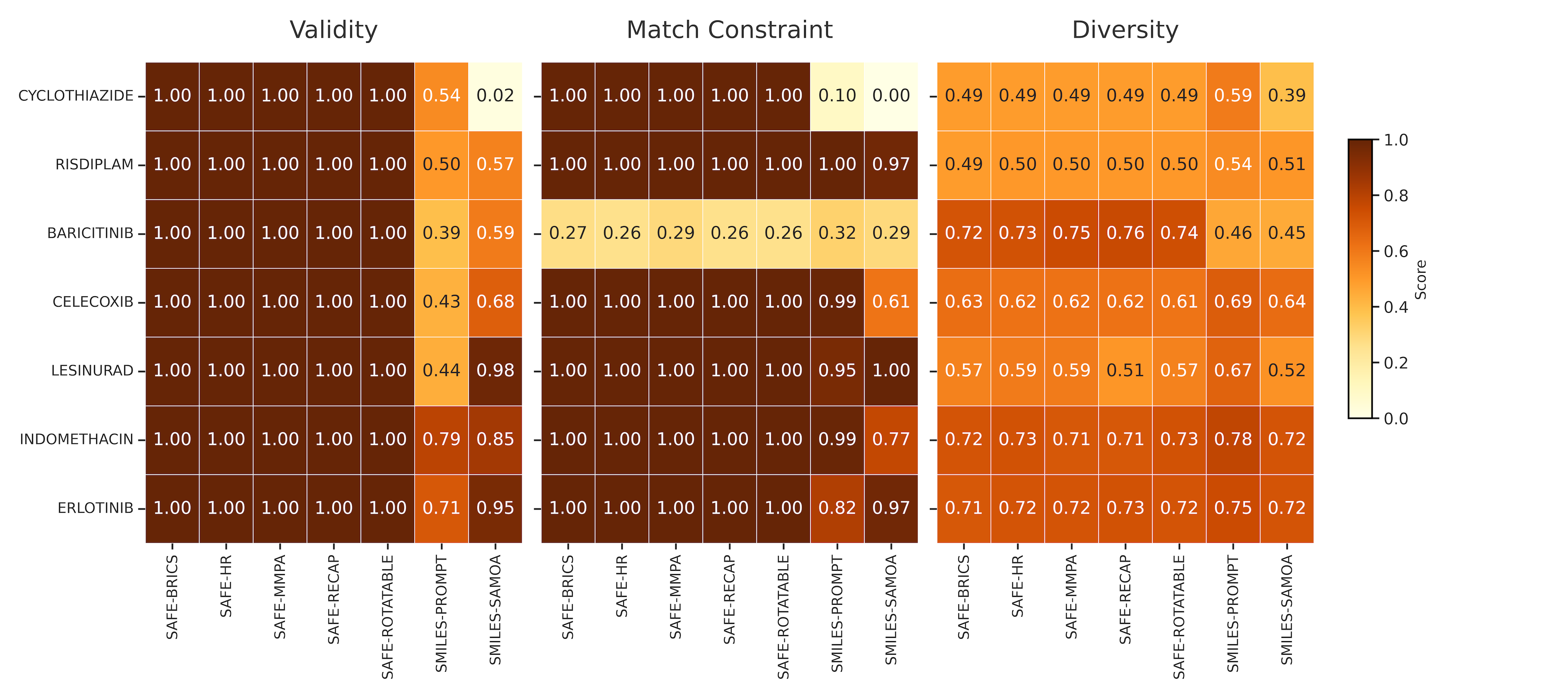}
    \caption{Per-drug performance of each fragment-constrained design method on the DRUG linker design benchmark.}
    \label{fig:per-linker-design}
\end{figure}

Figure~\ref{fig:per-benchmark-scaffold-dec} shows the per-dataset performance of each design method in scaffold decoration. Here, SAFE-based models outperform their SMILES counterparts, particularly in maintaining scaffold constraints. Figure~\ref{fig:per-benchmark-scaffold-dec-alt} visualizes the performance of each method on standard generative metrics, confirming the robustness of SAFE models. In Figure~\ref{fig:linker-design-alt}, we compare the performance of different fragment-constrained methods on the linker design task. SAFE-based methods achieve near-perfect validity and substructure preservation, outperforming SMILES-based approaches. The detailed breakdown in Figures~\ref{fig:per-scaffold-dec}~and~\ref{fig:per-linker-design} further highlights the strengths and limitations of each method.

\begin{figure}[h!]
    \centering
 \includegraphics[width=\linewidth]{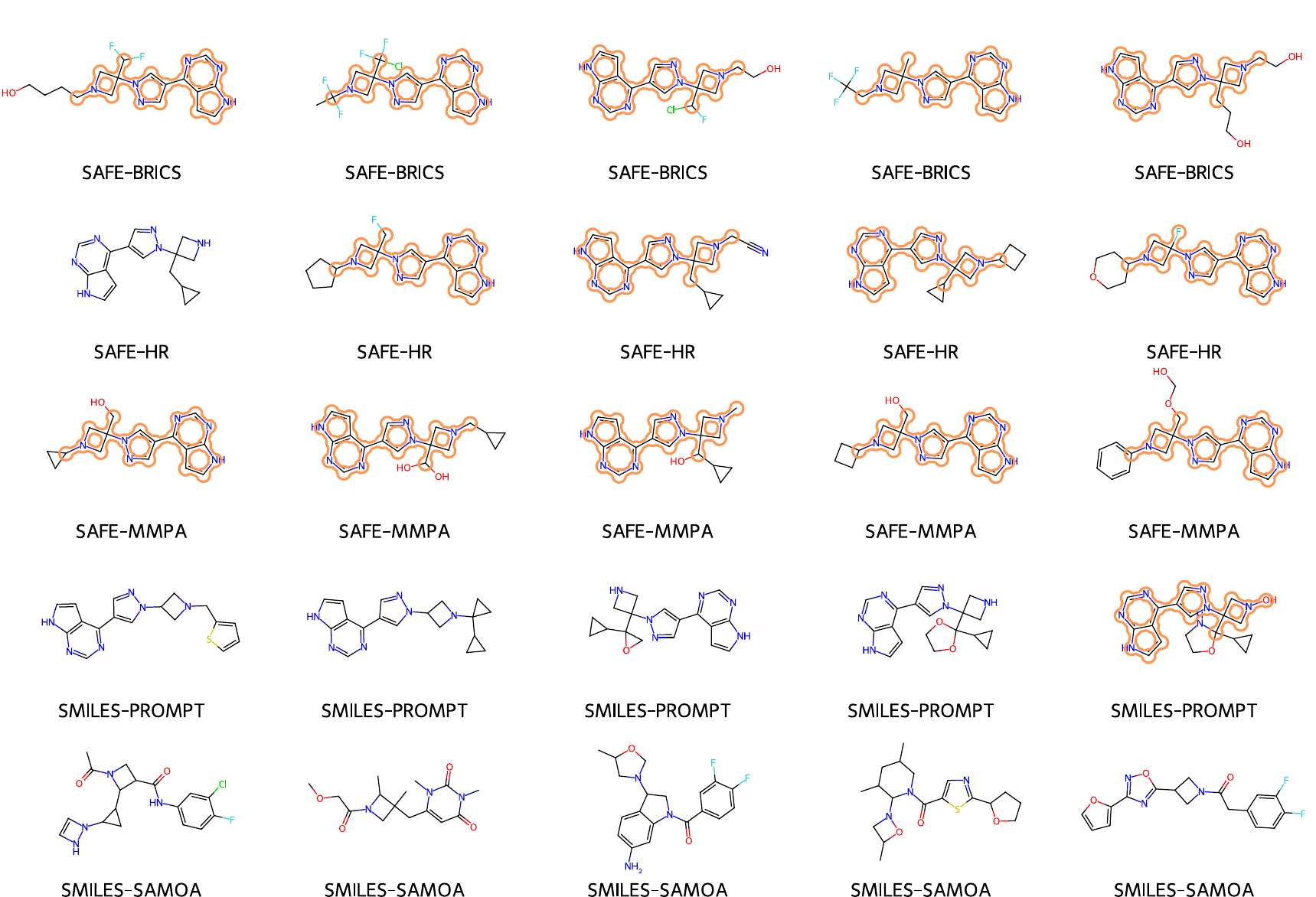}
    \caption{Example of sampled molecules under the fragment-constrained design of novel molecules sharing the same scaffold as the drug Baricitinib :\texttt{[*:1]N1CC([*:2])(n2cc(-c3ncnc4[nH]ccc34)cn2)C1}. The core being decorated is highlighted in the figure.}
    \label{fig:scaffold-dec-baricitinib}
\end{figure}

\begin{figure}[h!]
    \centering
 \includegraphics[width=\linewidth]{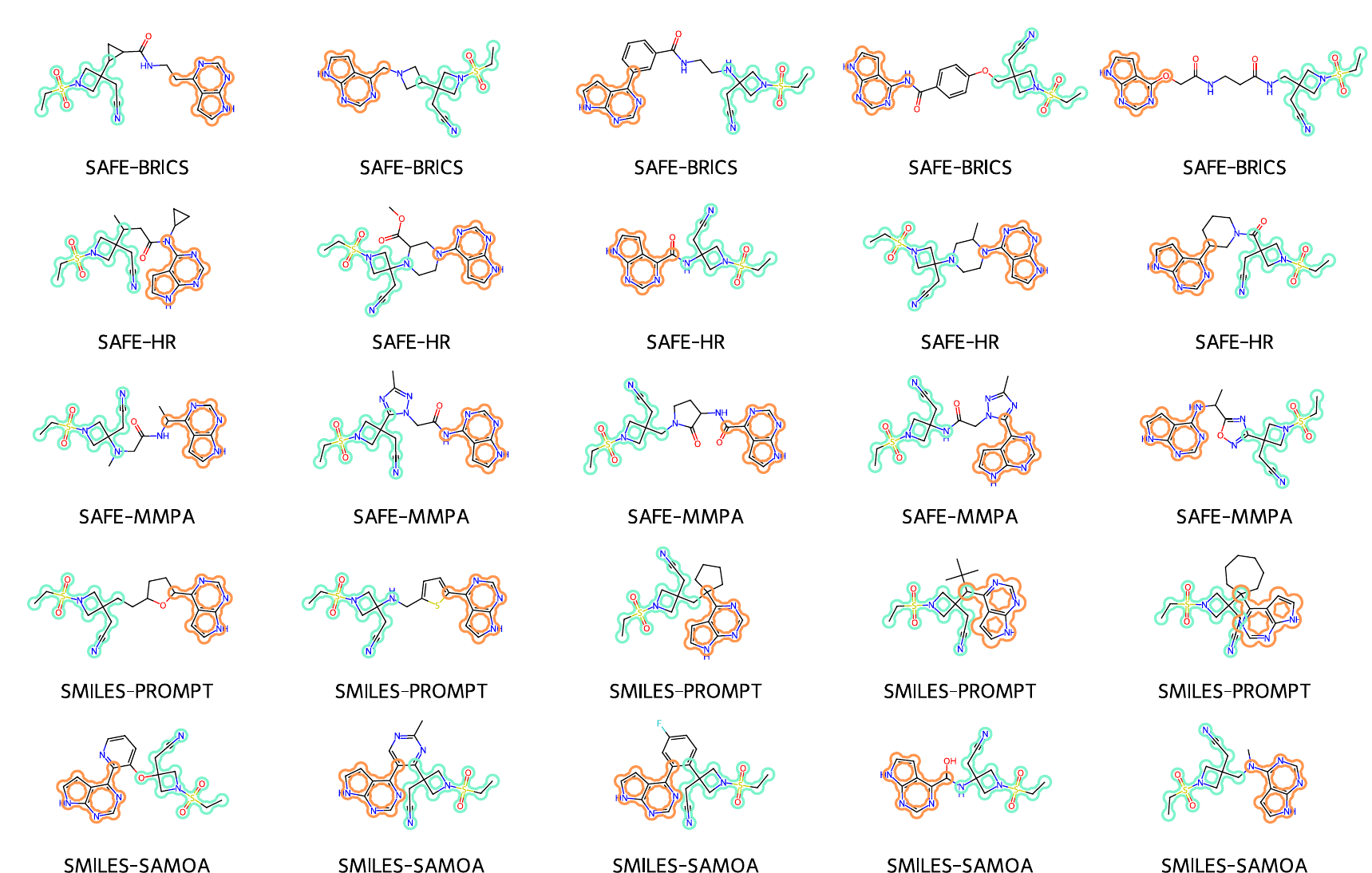}
    \caption{Example of sampled molecules under the fragment-constrained design of novel molecules by linking fragments of the drug Baricitinib: \texttt{[*:1]C1(CC\#N)CN(S(=O)(=O)CC)C1} and   \texttt{[*]c1ncnc2[nH]ccc12}. The two fragments linked are highlighted in the figure.}
    \label{fig:linker-design-baricitinib}
\end{figure}

\begin{figure}[h!]
    \centering
 \includegraphics[width=\linewidth]{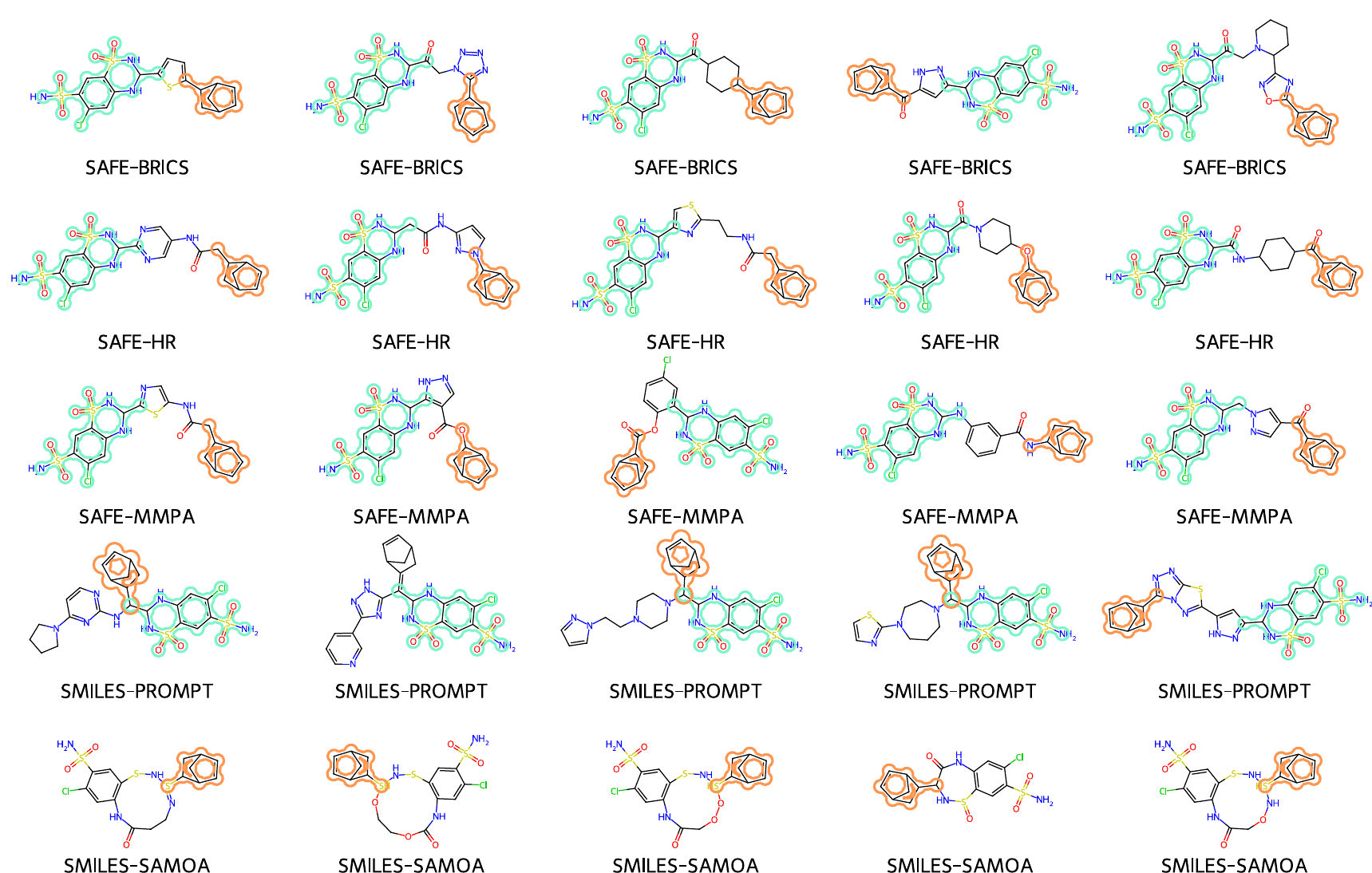}
    \caption{Example of sampled molecules under the fragment-constrained design of novel molecules by linking fragments of drug Cyclothiazide: \texttt{[*]C1CC2C=CC1C2}, \texttt{[*]C1Nc2cc(Cl)c(S(N)(=O)=O)cc2S(=O)(=O)N1}. The two fragments linked are highlighted in the figure.}
    \label{fig:linker-design-cyclothiazide}
\end{figure}

\clearpage

\end{document}